\pdfoutput=1
\documentclass[11pt]{article}

\usepackage{acl}

\usepackage{times}
\usepackage{latexsym}

\usepackage[T1]{fontenc}
\usepackage[utf8]{inputenc}

\usepackage{microtype}
\usepackage{graphicx}

\usepackage{multirow}
\usepackage{booktabs}
\usepackage{glossaries}
\newacronym[plural=LLMs]{llm}{LLM}{large language model}
\newacronym{amble}{AMBLE}{Archival Multi-task Benchmark for LLM Evaluation}
\newacronym{arcgpt}{ArcGPT}{Archival Generative Pre-trained Transformer}
\newacronym{post-ocr}{post-OCR}{post optical character recognition}

\title{ArcGPT: A Large Language Model Tailored for Real-world Archival Applications}

\author{\small Shitou Zhang\textsuperscript{1,4}, Jingrui Hou\textsuperscript{4,5}, Siyuan Peng\textsuperscript{1,4}, Zuchao Li\textsuperscript{2,4}, Qibiao Hu\textsuperscript{1,4}, Ping Wang\textsuperscript{1,3,4}\\
\small \textsuperscript{1}School of Information Management, Wuhan University;
\small  \textsuperscript{2}School of Computer Science, Wuhan University\\
\small  \textsuperscript{3}Center for the Studies of Information Resources, Wuhan University;
\small  \textsuperscript{4}Smart Archive Lab, Wuhan University\\
\small  \textsuperscript{5}Department of Computer Science, Loughborough University\\
\small \texttt{\{shitouzhang, pengsiyuan, zcli-charlie, huqibiao, wangping\}@whu.edu.cn}\\
\small \texttt{J.Hou@lboro.ac.uk}
}

\begin{document}
 \glsdisablehyper
\maketitle
\begin{abstract}

Archives play a crucial role in preserving information and knowledge, and the exponential growth of such data necessitates efficient and automated tools for managing and utilizing archive information resources. Archival applications involve managing massive data that are challenging to process and analyze. Although \glspl{llm} have made remarkable progress in diverse domains, there are no publicly available archives tailored \glspl{llm}. Addressing this gap, we introduce {\bf \gls{arcgpt}}, to our knowledge, the first general-purpose \gls{llm} tailored to the archival field. To enhance model performance on real-world archival tasks, \gls{arcgpt} has been pre-trained on massive and extensive archival domain data. Alongside \gls{arcgpt}, we release \gls{amble}, a benchmark comprising four real-world archival tasks. Evaluation on \gls{amble} shows that \gls{arcgpt} outperforms existing state-of-the-art models, marking a substantial step forward in effective archival data management. Ultimately, \gls{arcgpt} aims to better serve the archival community, aiding archivists in their crucial role of preserving and harnessing our collective information and knowledge.

\end{abstract}

\section{Introduction}
\glsreset{arcgpt}
\glsreset{amble}
\glsreset{llm}
Archival resources encompass a diverse range of evidential materials, recollections, and repositories of knowledge meticulously curated by their creators to safeguard their legitimate rights and interests within a specific context \cite{williams2002trusting,an2014reinventing,henninger2016new,an2017knowledge}. Archiving electronic records has witnessed significant advancements worldwide over the last two decades. For instance, in China, archival resources hold high value for the government and numerous archival institutions \cite{moss1996dang,xiao2021security} have been established to manage and preserve these resources \cite{an2017knowledge}. By the end of 2021, the total volume of electronic archives in China reached an impressive 1629.9TB \cite{saac2022}. These invaluable resources serve a multifaceted purpose, facilitating the identification, evaluation, preservation, and accessibility of documentary materials of enduring significance to the broader public \cite{roper2003archives}. Furthermore, their existence plays a vital role in assessing the accountability of various institutions, as they diligently preserve and provide access to public records in accordance with legal and ethical principles. From a national perspective, the utilization of these archival resources assumes paramount significance, particularly in fostering collaborative knowledge services that harness the potential of archival materials as essential societal knowledge assets \cite{an2017knowledge}.

Given the substantial volume and significant value of archival resources, there is an urgent need to deploy automatic and intelligent tools to process digital archives and replace labor-intensive manual methods \cite{moss2018reconfiguration,aangenendt2022archives}. To date, a considerable number of artificial intelligence methods have been applied in this field, including archival appraisal \cite{hutchinson2020natural, shabou2020algorithmic}, sensitive information identification \cite{hutchinson2018protecting}, archival epoch classification \cite{blanke2017identifying}, and archival retrieval \cite{lee2019machine,lansdall2020history}. Despite the progress made by these methods and applications in the basic groundwork of archival resource processing and management, more advanced tasks, such as archival data understanding and reasoning, still represent a significant research gap. With the current prevalence of \glspl{llm}, which are capable of extracting and understanding intent information from human-provided instructions \cite{cao2023comprehensive}, this research aims to address the aforementioned research gap by developing a new \gls{llm}, specifically tailored for the archival domain.

Our recently developed \gls{llm} tailored for real-world archival applications is named \gls{arcgpt}. Building upon the BatGPT architecture \cite{li2023batgpt},  \gls{arcgpt} underwent extensive training on vast archival domain data. This comprehensive training exposed the model to historical language usage, specialized jargon, and context-specific knowledge, endowing  \gls{arcgpt} with the capability to proficiently interpret and process archival data, a challenge often encountered by generic language models.

The rising demand from archival workers and institutions prompted the introduction of \gls{amble}. This benchmark serves as a thorough evaluation framework for assessing the performance of language models on real-world archival tasks. \gls{amble} encompasses data from four distinct tasks: retention period prediction, open access identification, confidentiality prediction, and\gls{post-ocr} processing. By incorporating these tasks, \gls{amble} provides a robust and specialized framework to gauge the effectiveness and versatility of \glspl{llm} in archival domains.

In the evaluation conducted on \gls{amble}, \gls{arcgpt} exhibited superior performance compared to existing generic \glspl{llm}. This noteworthy advancement represents a significant stride in the realm of automated archival data management and utilization. We firmly believe that  \gls{arcgpt} and \gls{amble} will form the bedrock for future research and development in this crucial and relatively unexplored domain.

\section{Related Work}

LLMs refer to pre-trained language models that contain hundreds of billions (or more) of parameters, which are trained on massive text data \cite{Shanahan2022TalkingAL}. They exhibit outstanding performance in various natural language processing (NLU) tasks and domains, and therefore have attracted considerable attention \cite{Fan2023ABR}. Moreover, as the capacities of LLMs continue to advance, benchmarks play a crucial role in evaluating their development \cite{li2023cmmlu}. Existing studies on LLMs and their evaluation benchmarks can be divided into two categories based on their application domain: general LLMs and general-purpose LLM evaluation benchmarks, and domain-specific LLMs and domain-specific LLM evaluation benchmarks. In this section, we summarize the relevant research on both general and domain-specific LLMs and their evaluation benchmarks, and discuss the necessity and potential of developing a LLM and evaluation benchmark for the archival domain.

\subsection{General LLMs and General-purpose Evaluation Benchmarks}

General LLMs are LLMs that are trained on datasets encompassing a wide range of topics and domains. They have been shown to be effective on a variety of language-related tasks with general-purpose capabilities, and are posing a significant impact on the AI community \cite{Zhao2023ASO}. As per recent research, some of the top general LLMs have been announced and released in the last few years. In 2020, the release of GPT-3 \cite{Brown2020LanguageMA} by OpenAI exemplified the significant benefits of training LLMs and propelled the field of NLU forward. GPT-3 has 175 billion parameters, a hundredfold increase over the previous GPT-2 model, and did remarkably well across a wide range of main LLM tasks. Following the success of GPT-3, several other general LLMs have emerged. PaLM \cite{Chowdhery2022PaLMSL}, a 540-billion parameter, densely activated, Transformer language model, was introduced in 2022. It has strong capabilities in multilingual tasks and source code generation, which was demonstrated on a wide array of benchmarks. GLM \cite{Zeng2022GLM130BAO}, a bilingual (English and Chinese) pre-trained language model with 130 billion parameters, was released in 2022. The resulting GLM-130B model significantly outperforms GPT-3 175B on various popular English benchmarks. BLOOM \cite{Scao2022BLOOMA1} is an open-access language model with 176 billion parameters that was trained on the ROOTS corpus. It has been found that BLOOM achieves competitive performance on a wide range of benchmarks, with even stronger results after undergoing multitask-prompted finetuning.  LLaMA \cite{Touvron2023LLaMAOA} is a set of foundational language models with a parameter range of 7 billion to 65 billion, trained on trillions of tokens. Experimental results show that LLaMA-13B outperforms GPT-3 (175B) on most benchmarks, while LLaMA-65B is competitive with the best models, Chinchilla-70B and PaLM-540B. In 2023, GPT-4 \cite{OpenAI2023GPT4TR}, a large-scale, multimodal language model capable of accepting both image and text inputs and producing text outputs, was developed. This Transformer-based model is pre-trained to predict the next token in a document and has achieved human-level performance on various professional and academic benchmarks. Although general LLMs have a broad range of capabilities in performing language-related tasks, they have limitations in implementing NLU on the archival text and performing archival domain tasks due to archival domain-specific complexity and terminology.

To fully evaluate the overall performance of general LLMs in NLU tasks, various general-purpose LLM evaluation benchmarks have been proposed, such as SentEval, GLUE, and Super-GLUE. To be specific, SentEval \cite{conneau2018senteval} is a toolkit specifically designed to evaluate the quality of universal sentence representations. It offers a diverse set of tasks, including binary and multi-class classification, natural language inference, and sentence similarity, among others. By encompassing a broad spectrum of tasks, SentEval provides a comprehensive evaluation of the generalization ability of sentence representation models and allows for a fair comparison of different models. GLUE \cite{wang2019glue} is a tool that serves to evaluate and analyze the performance of models across a diverse range of NLU tasks. It is designed to be model-agnostic, meaning that it can be used to evaluate the performance of any NLU model, regardless of its architecture. GLUE encompasses a wide range of tasks that include sentiment analysis, question answering, and natural language inference, among others. Super-GLUE \cite{wang2020superglue} is a new benchmark that builds upon the GLUE with a new set of more difficult language understanding tasks, a software toolkit, and a public leaderboard. While these benchmarks have made significant progress in evaluating NLU tasks, their primary focus has been on assessing language skills. As a result, they have become less commonly used as benchmarks for LLMs, as many of these models are now capable of generating fluent and plausible language \cite{li2023cmmlu}. Meanwhile, various benchmarks have been proposed to evaluate LLMs’ performance in different aspects, including question answering \cite{Rajpurkar2018KnowWY, Kwiatkowski2019NaturalQA, Li2022MultiSpanQAAD}, knowledge reasoning \cite{Clark2018ThinkYH, Talmor2019CommonsenseQAAQ, sawada2023arb}, and code generation \cite{Chen2021EvaluatingLL, Austin2021ProgramSW}. While general-purpose evaluation benchmarks have been instrumental in evaluating the overall language capabilities of LLMs, they may not be able to capture the nuances and complexities of specific domains. As a result, these benchmarks may have limitations when it comes to assessing the performance of LLMs in specific domains.

\subsection{Domain-specific LLMs and Domain-specific Evaluation Benchmarks}

Due to the limitations of general LLMs in handling specific domain tasks, some researchers have developed domain-specific LLMs that are trained on texts specific to a particular domain. The aim of domain-specific LLMs is to capture domain knowledge, terminology, and style, and to improve the performance of various downstream tasks in that domain \cite{wang2023mediagpt}. Currently, researchers are mainly focused on yielding LLMs for the domains of finance, medicine, and science \cite{Pahune2023SeveralCO}. These models have revealed the advantages of building domain-specific LLMs. Specifically, regarding the LLMs for the financial domain, \citet{Wu2023BloombergGPTAL} developed the first financial LLM BloombergGPT, a 50 billion parameter language model that is specifically designed to support various tasks within the financial industry. To train the model, they constructed a massive dataset consisting of 363 billion tokens, which includes data from Bloomberg's extensive sources as well as 345 billion tokens from general-purpose datasets. \citet{Xie2023PIXIUAL} proposed the financial domain-specific LLM FinMA by conducting the multi-task instruction tuning on LLaMA with the building dataset. Experimental results showed that FinMA significantly outperforms LLMs, including BloombergGPT, ChatGPT, and GPT-4 on most tasks in the financial domain. In the medical domain, \citet{Xiong2023DoctorGLMFY} developed a healthcare-specific LLM named DoctorGLM by utilizing the ChatGLM-6B model \cite{Du2021GLMGL}. DoctorGLM is an open-source, bilingual language model based on the GLM framework with 6.2 billion parameters. Additionally, \citet{Wu2023PMCLLaMAFF} introduced PMC-LLaMA, an open-source language model that is fine-tuned on a total of 4.8 million biomedical academic papers to further incorporate medical knowledge and enhance its capability in the medical domain. Furthermore, in the science domain,\citet{Taylor2022GalacticaAL} developed Galactica, a LLM that can store, combine, and reason about scientific knowledge based on a large corpus of scientific papers, reference materials, knowledge bases, and other sources. These studies have demonstrated the importance of tailoring the LLMs specifically for the specific domain, which motivates further development of models focused on specific domains. However, it was found that there are no LLMs for the archival domain, which is not conducive to the accomplishment of the archival domain’s tasks.

To evaluate the performance of LLMs for specific domains, researchers have constructed a number of domain-specific evaluation benchmarks. After reviewing relevant literature, it has been observed that domain-specific evaluation benchmarks are predominantly concentrated in the financial and medical domains. In terms of financial Evaluation Benchmark, \citet{Shah2022WhenFM} developed FLUE, a comprehensive suite of open-source benchmarks for the financial domain. FLUE includes five new benchmarks for various NLP tasks related to finance, as well as commonly used benchmarks from previous research. \citet{chen2022convfinqa} proposed a new large-scale dataset, ConvFinQA, aiming to study the chain of numerical reasoning in conversational finance question answering. \citet{Xie2023PIXIUAL} built the FLARE Benchmark covering 4 financial NLP tasks with 6 datasets, and 1 financial prediction task with 3 datasets to evaluate the proposed model FinMA and other LLMs holistically. \citet{Lu2023BBTFinCC} proposed BBT-CFLEB, a Chinese Financial Language understanding and generation Evaluation Benchmark, which includes six datasets covering both understanding and generation tasks. In the medical domain, \citet{jin2019pubmedqa} constructed PubMedQA, the first biomedical question-answering dataset collected from PubMed abstracts. \citet{pal2022medmcqa} proposed MedMCQA, a new large-scale, multiple-choice question-answering dataset designed to address real-world medical entrance exam questions. These specific-domain evaluation benchmarks are crucial for assessing the performance of general LLMs on specific domains or the performance of specific-domain LLMs. This is important for promoting the development of LLMs. However, there is currently a lack of evaluation benchmark datasets in the archival domain. This hinders the application of LLMs in the archival domain and the development of LLMs specifically tailored for this domain.

In summary, due to the specialized and complex nature of textual data in the archival domain, existing general LLMs have limitations when performing language-related tasks in this area, but no LLMs have been developed specifically for the archival domain yet. Additionally, in terms of evaluation benchmarks, researchers have mainly constructed general-purpose benchmarks to assess the NLU performance of LLMs, but these benchmarks have limitations when evaluating the performance of LLMs in the archival domain and there are no evaluation benchmarks tailored for different archival tasks. Therefore, this paper aims to develop an LLM specifically for the archival domain, called ArcGPT, and release a new benchmark called AMBLE, which includes four major archival tasks to aid in evaluating LLMs in the archival domain.

\section{ArcGPT}
\glsreset{arcgpt}
\gls{arcgpt} is a 7B \gls{llm} specifically tailored for archival applications. It inherits its foundational structure from the BatGPT architecture \cite{li2023batgpt}, which has already shown remarkable efficiency in various natural language tasks. 

The uniqueness of \gls{arcgpt} lies in its pretraining process on extensive archival domain data. This data includes a diverse array of document types and styles, from archival journals to archival records. Additionally, the data spans numerous historical periods, exposing the model to the evolution of language usage, phraseology, and context-specific knowledge. The model's expansive training on this specialized data endows it with a robust understanding of archival language nuances and the ability to interpret complex historical contexts that generic language models might find challenging.

The impact of the domain-specific pretraining on \gls{arcgpt}'s performance is palpable. Not only does it enhance the model's understanding of archival terminologies and historical language constructs, but it also improves its ability to process and analyze the context-rich, diverse data often encountered in archives. For instance, when tested on tasks such as document classification, \gls{arcgpt} showed the superior capability to identify relevant themes and patterns across diverse data types and historical periods. Similarly, the model exhibited significant improvements in the post-OCR processing task where the model is required to correct noisy OCR output, underlining its comprehensive understanding of language intricacies in long-range and complex contexts.

\section{AMBLE}
\glsreset{post-ocr}

\subsection{Task Overview}
AMBLE encompasses four tasks integral to modern archival work: retention period prediction, open-access identification, confidentiality prediction, and post-OCR processing. The retention period prediction task involves estimating the period for which documents need to be preserved based on their content and relevance. This task is critical in archival management as it influences the allocation of resources and storage space. open access identification, the second task, involves determining whether a document can be made publicly accessible or not, taking into account factors such as confidentiality, security, and legal obligations. The third task, confidentiality prediction, requires the model to predict whether a document contains sensitive information based on its content. This task is especially crucial given the heightened need for privacy and data protection. Finally, post-OCR processing involves cleaning and correcting noisy OCR-processed texts, a task essential for converting scanned archival documents into machine-readable text with a high level of accuracy.

\subsection{Data Annotation and Format}

Archives serve as valuable corporate and institutional assets, playing a crucial role in preserving historical records and important information. However, the acquisition channels of these documents are often undisclosed by their respective owners, which poses significant challenges in training large language models in the field of archives. To address this issue, we took the initiative to collaborate with the archives agency of a specific administrative unit in China. Through this collaboration, we were able to obtain a substantial collection of archival documents, directly sourced from authentic administrative activities.

The archival documents procured for our study were in electronic format, having undergone a scanning process to digitize paper-based archives. These electronic scans preserved essential information such as "subject", "source organization", "formation time", "archiving time" and "textual content". However, some of information was stored in the form of images, creating a challenge for direct text-based analysis. To mitigate this impediment, we employed OCR technology to extract the textual information from the images. It is worth mentioning that the OCR process introduced some degree of noise data, which we took into account during our analysis.

To ensure accurate labeling of the acquired archives for our research, we engaged students specializing in archival science. Their expertise and understanding of archival principles proved invaluable in accurately annotating the data. Recognizing their contribution, we provided reasonable remuneration for their efforts. The resulting dataset, meticulously labeled by these students, forms the foundation of our research and is presented in Table~\ref{table-data-stats}.

\begin{figure}[h]
    \centering
    \includegraphics[width=0.45\textwidth]{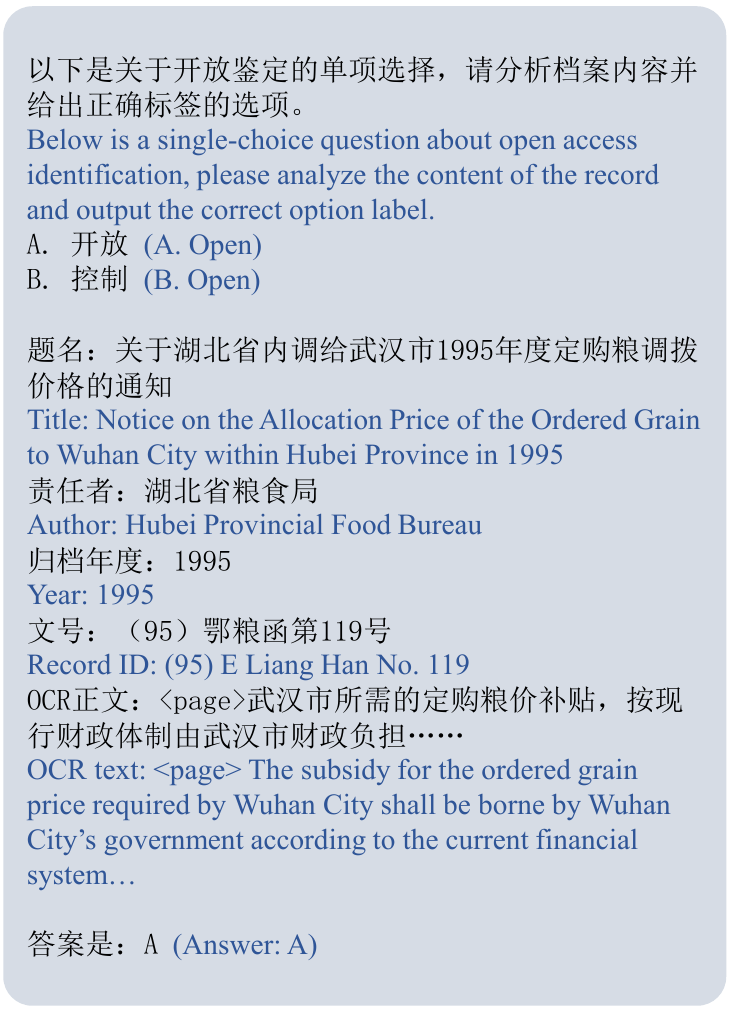}
    \caption{Prompted example from AMBLE.}
    \label{figure-input-format}
\end{figure}

To evaluate LLMs performance, we wrap the data examples in AMBLE into prompted instructions using task-specific templates. One example is presented in Figure~\ref{figure-input-format}. The prompted instruction encapsulates crucial document attributes such as "titile", "source organization", "formation time", and "record ID".

\begin{table}[htbp]
\centering
\begin{tabular}{lrr}
\toprule
Task                        & \multicolumn{1}{c}{\#Train} & \multicolumn{1}{c}{\#Test} \\ \midrule
Retention Period Prediction & 2,771                        & 250                        \\
Open Access Identification  & 7,002                        & 250                        \\
Confidentiality Prediction  & 3,579                        & 250                        \\
Post-OCR Processing         & 1,117                        & 250                        \\ \midrule
Sum                         & 14,469                       & 1,000                       \\
\bottomrule
\end{tabular}
\caption{Statistics of the AMBLE dataset.}
\label{table-data-stats}
\end{table}

\section{Evaluation}

\subsection{Baseline Models}

In order to assess the effectiveness of \gls{arcgpt} in comparison to other models on the AMBLE benchmark, we measured its performance against a variety of state-of-the-art models. For classification tasks, the following baseline models are included:

 \begin{itemize}
    \item BERT-wwm-ext \cite{cui-etal-2020-revisiting}: BERT-wwm-ext builds upon BERT's architecture and incorporates specific improvements tailored for Chinese language understanding tasks. It is widely utilized in various Chinese natural language processing applications.

    \item RoBERTa-wwm-ext\cite{cui-etal-2020-revisiting}: RoBERTa-wwm-ext is an extended and refined version of the RoBERTa \cite{liu2019roberta} model. It has emerged as a prominent choice for pre-trained language models in Chinese NLP tasks due to its robust language representation capabilities.  
    
    \item ChatGLM-6B \cite{Zeng2022GLM130BAO}: a Chinese-English bilingual large language model with roughly 6 billion parameters. Similar to ChatGPT, ChatGLM-6B is specifically optimized for Chinese question-and-answer (Q\&A) scenarios and dialogue interactions.

    \item Chinese-LLaMA-Alpaca \cite{cui2023efficient}: This model builds upon the original LLaMA-7B \cite{touvron2023llama} and utilizes Alpaca-like \cite{alpaca} instruction tuning to further enhance instruction understanding. It expands the Chinese vocabulary and incorporates Chinese data for secondary pre-training.
\end{itemize}
For the post-OCR processing task, an additional pair of generative models were evaluated alongside ChatGLM-6B and Chinese-LLaMA-Alpaca:

\begin{itemize}
    \item BART-Large-csc \cite{shao2021cpt}: This variant of BART-Large \cite{lewis2020bart} has been pre-trained on a substantial corpus of Chinese text and specialized for Chinese Spelling Correction (CSC).

    \item Mengzi-T5-Base-csc \cite{zhang2021mengzi}: This model builds on the T5 architecture \cite{raffel2020exploring}, and was pre-trained on a massive 300 GB corpus of Chinese text.
\end{itemize}
It's worth noting that before their evaluation on AMBLE, the two CSC models had been fine-tuned on the SIGHAN \cite{tseng-etal-2015-introduction} and Wang271K \cite{wang-etal-2019-confusionset} datasets, equipping them with a robust ability to handle a wide variety of spelling errors in Chinese language.

\subsection{Results}

\subsubsection{Retention period prediction, open-access identification, and confidentiality prediction}

We first evaluate models on the three classification tasks: retention period prediction, open-access identification, and confidentiality prediction. To assess the overall performance of both baseline models and \gls{arcgpt}, we employed precision, recall, and F1 score as evaluation metrics. The results of these evaluations are presented in Table~\ref{table-clf-results}. Notably, \gls{arcgpt} demonstrated superior F1 scores of 84.40, 84.00, and 94.4 for open-access identification, retention period prediction, and confidentiality prediction, respectively. These scores surpassed those achieved by the other two LLM baselines. In comparison, the predictive model RoBERTa-wwm-ext exhibited the highest F1 scores among all models tested, attaining values of 88.80, 88.00, and 97.20 for the three classification tasks in \gls{amble}, respectively.

Although the proposed \gls{arcgpt} showed remarkable performance and outperformed other strong generative baseline models in the three classification tasks in \gls{amble}, it is crucial to acknowledge that predictive models possess inherent differences from generative models. Consequently, despite its success, \gls{arcgpt} still exhibits a notable performance gap when compared to the best predictive models. Further research and refinement of generative approaches may be required to bridge this gap and attain even more competitive results.

\begin{table*}[htbp]
\centering
\begin{tabular}{lccccccccccc}
\toprule
\multirow{2}{*}{Model} & \multicolumn{3}{c}{\textbf{Retention Period}}                                & \multicolumn{1}{c}{\textbf{}} & \multicolumn{3}{c}{\textbf{Open Access}}                                     &  & \multicolumn{3}{c}{\textbf{Confidentiality}}                                 \\ \cline{2-4} \cline{6-8} \cline{10-12} 
                       & \multicolumn{1}{c}{Prec} & \multicolumn{1}{c}{Rec} & \multicolumn{1}{c}{F-1} & \multicolumn{1}{c}{}          & \multicolumn{1}{c}{Prec} & \multicolumn{1}{c}{Rec} & \multicolumn{1}{c}{F-1} &  & \multicolumn{1}{c}{Prec} & \multicolumn{1}{c}{Rec} & \multicolumn{1}{c}{F-1} \\ \midrule
BERT-wwm-ext           & 71.92                    & 95.42                   & 74.40                   &                               & 77.94                    & 82.17                   & 78.80                   &  & 97.51                    & 97.03                   & 95.60                  \\
RoBERTa-wwm-ext        & 69.46                    & 92.16                   & 70.40                   &                               & 78.26                    & 83.72                   & 79.60                   &  & 98.02                    & 98.02                   & 96.80                  \\
ChatGLM                & 80.46                    & 91.50                   & 81.20                   &                               & 83.47                    & 78.29                   & 80.80                   &  & 88.94                    & 99.50                   & 89.60                  \\
Chinese-LLaMA-Alpace   & 57.21                    & 83.01                   & 51.60                   &                               & 52.38                    & 93.80                   & 52.80                   &  & 93.92                    & 84.16                   & 82.80                  \\
ArcGPT                 & 93.18                    & 80.39                   & 84.40                   &                               & 83.97                    & 85.27                   & 84.00                   &  & 94.76                    & 98.51                   & 94.40                  \\ \bottomrule
\end{tabular}
\caption{Evaluation results of the three classification tasks.}
\label{table-clf-results}
\end{table*}

\subsubsection{Post-OCR processing}

The results of  \gls{post-ocr} task in \gls{amble}, including \gls{arcgpt} and other baseline models, are presented in Table~\ref{table-seq-results}. Remarkably, Bart-Large-csc and Mengzi-T5-Base-csc, both trained on extensive Chinese spelling error datasets, exhibit significantly superior performance compared to the other models. Notably, Mengzi-T5-Base-csc achieves the highest performance, with an impressive Levenshtein Distance of 10.90. \gls{arcgpt}'s performance closely resembles that of ChatGLM, with Levenshtein Distances of 38.86 and 37.41, respectively, slightly surpassing Chinese-LLaMA-Alpaca.

The outcomes of this experiment underscore the advantages of pre-training on specific misspelling correction datasets, as evidenced by the superior performance of Bart-Large-csc and Mengzi-T5-Base-csc. Conversely, the other three models exhibit much lower performance in comparison. \gls{arcgpt} fails to demonstrate a substantial advantage over the other models, necessitating dedicated efforts to enhance its performance in future endeavors. 

\begin{table}[htbp]
\centering
\begin{tabular}{lr}
\toprule
Model                & \multicolumn{1}{c}{Levenshtein Distance} \\ \midrule
Bart-Large-csc       & 18.50                                    \\
Mengzi-T5-Base-csc   & 10.90                                    \\
ChatGLM              & 37.41                                    \\
Chinese-LLaMA-Alpaca & 41.94                                        \\
ArcGPT               & 38.86                                    \\
\bottomrule
\end{tabular}
\caption{Model performance on the Post-OCR processing task.}
\label{table-seq-results}
\end{table}

\section{Future Work and Conclusion}

Archival data holds a significant responsibility in preserving information and knowledge. The ever-increasing volume of archive data has highlighted the need for efficient and automated tools that can enhance the management and utilization efficiency of archive information resources while also alleviating the workload of archivists. The availability of vast archive data has opened up new opportunities for training large language models specialized in the archival domain.

In light of this context, this paper introduces \gls{arcgpt}, the first-ever general-purpose large language model tailored for the archives field. Additionally, it presents the \gls{amble} benchmark, comprising four real-world archive tasks. \gls{arcgpt} has exhibited remarkable performance on three classification tasks in the \gls{amble} benchmark, surpassing the current state-of-the-art LLMs. However, it is essential to acknowledge that some gaps still exist when comparing \gls{arcgpt} with predictive models. Furthermore, it is evident that \gls{arcgpt}'s performance in the post-OCR task is subpar, indicating a clear area for improvement. Addressing and enhancing the model's performance in this specific task constitutes a primary focus of our ongoing research efforts. 

The performance of \gls{arcgpt} showcases the tremendous potential of large language models in the archival domain. Moving forward, we are committed to further optimizing \gls{arcgpt} to enhance its capabilities. We also encourage archivists to embrace the use of \gls{arcgpt} in their daily work and actively provide feedback and suggestions. This collaborative approach will enable \gls{arcgpt} to better cater to the needs of archivists and archives, ultimately advancing the field and its valuable contributions to preserving our collective history and knowledge.

% Entries for the entire Anthology, followed by custom entries
\bibliography{custom}
\bibliographystyle{acl_natbib}

\end{document}